\documentclass[aps,prl,reprint, superscriptaddress]{revtex4-2}

\usepackage[utf8]{inputenc} 
\usepackage[T1]{fontenc}    
\usepackage{lmodern}
\usepackage{amsfonts}       
\usepackage{amsmath}
\usepackage{amsbsy}

\usepackage{graphicx}
\graphicspath{{fig/}}

\begin{document}

\title{Black and Gray Box Learning of Amplitude Equations:\\ Application to Phase Field Systems}

\author{Felix P. Kemeth}
\affiliation{
  Department of Chemical and Biomolecular Engineering,
  Whiting School of Engineering, Johns Hopkins University,
  3400 North Charles Street, Baltimore, MD 21218, USA
}
\author{Sergio Alonso}
\affiliation{
  Universitat Polit\`{e}cnica de Catalunya (UPC),
  Department of Physics,
  C. Jordi Girona, 1-3 08034 Barcelona, Spain
}
\author{Blas Echebarria}
\affiliation{
  Universitat Polit\`{e}cnica de Catalunya (UPC),
  Department of Physics,
  C. Jordi Girona, 1-3 08034 Barcelona, Spain
}

\author{Ted Moldenhawer}
\affiliation{
  Universit\"{a}t Potsdam,
  Institut f\"{u}r Physik und Astronomie,
  Karl-Liebknecht-Str. 24/25, D-14476 Potsdam-Golm, Germany
}

\author{Carsten Beta}
\affiliation{
  Universit\"{a}t Potsdam,
  Institut f\"{u}r Physik und Astronomie,
  Karl-Liebknecht-Str. 24/25, D-14476 Potsdam-Golm, Germany
}

\author{Ioannis G. Kevrekidis}
\email[]{yannisk@jhu.edu}
\affiliation{
  Department of Chemical and Biomolecular Engineering,
  Whiting School of Engineering, Johns Hopkins University,
  3400 North Charles Street, Baltimore, MD 21218, USA
}

\date{\today}

\begin{abstract}
  We present a data-driven approach to learning surrogate models for amplitude equations, and illustrate its application to interfacial dynamics of phase field systems.
  In particular, we demonstrate learning effective partial differential equations describing the evolution of phase field interfaces from full phase field data.
  We illustrate this on a model phase field system, where analytical approximate equations for the dynamics of the phase field interface (a higher order eikonal equation and its approximation, the Kardar–Parisi–Zhang (KPZ) equation) are known.
  For this system, we discuss data-driven approaches for the identification of equations that accurately describe the front interface dynamics.
  When the analytical approximate models mentioned above become inaccurate, as we move beyond the region of validity of the underlying assumptions, the
  data-driven equations outperform them. In these regimes, going beyond black-box identification, we explore different approaches to learn data-driven corrections to the analytically approximate models, leading to effective gray box partial differential equations.
\end{abstract}

\maketitle

\section{Introduction}

Phase field models provide an effective mathematical framework to investigate the 
dynamics of interfacial boundaries~\cite{04_free_bound_probl}.
They have been successfully used in fields such as material science,
to describe phase separation~\cite{cahn58_free_energ_nonun_system} and melting processes in alloys~\cite{allen72_groun_state_struc_order_binar, allen76_mechan_phase_trans_within_miscib, allen79_micros_theor_antip_bound_motion}, the formation of microstructures~\cite{li17_review}, solidification~\cite{boettinger02_phase_field_simul_solid}, solute precipitation~\cite{xu08_phase_field_model_solut_precip_dissol}, crack propagation~\cite{miehe10_phase_field_model_rate_indep_crack_propag} or to model grain growth~\cite{suwa05_comput_simul_grain_growt_three, suwa05_phase_field_simul_effec_anisot},
as well as in biomechanics, for example to model the development of fractures~\cite{raina15_phase_field_model_fract_biolog_tissues}, cell migration~\cite{shao_computational_2010,aranson_physical_2016,alonso18_model_random_crawl_membr_defor,moreno20_modeling_cell_crawling,moure17_phase_field_model_isogeom_analy_cell_crawl, najem16_phase_field_model_collec_cell_migrat}, tumor growth~\cite{lima14_analy_numer_solut_stoch_phase}, dendrite growth~\cite{takaki14_phase_field_model_simul_dendr_growt}, and mechanotransduction~\cite{iskratsch14_apprec_force_shape_rise_mechan_cell_biolog}, to name a few.
See Ref.~\cite{gomez19_review_comput_model_phase_trans_probl} for a recent review.
Such phase field models typically rely on a phase $\phi$, ranging from -1 to 1 (or 0 to 1). One then classifies a part of the domain where $\phi \approx -1$ as one state,
whereas the part where $\phi \approx 1$ corresponds to a different state,
both phases being separated by an interface, where the phase $\phi$ transitions from one state to the other.\\

Enormous effort has been invested into deriving evolution equations for the position of such interfaces; a prominent example involves the reduction of the dynamics of binary alloys~\cite{caginalp98_analy_phase_field_alloy_trans_layer}.
The same is true for the derivation of equations like the Kuramoto-Sivashinsky equation, describing the height of a thin water film flowing down an inclined surface~\cite{kuramoto78_diffus_induc_chaos_react_system, sivashinsky77_nonlin_analy_hydrod_instab_lamin_flames_i}, thereby reducing the dynamics of a free boundary problem to an amplitude equation.
Envelope equations approximating the long-wavelength motions of fluids or of Fermi-Pasta-Ulam-Tsingou solutions~\cite{schneider00_count_waves_fluid_surfac_contin, schneider10_bound_nonlin_schroed_approx_fermi}
also fall under the class of systems, where the techniques described here are applicable.
%
Having access to effective interface (or amplitude, or modulation envelope) equations does not only significantly reduce the computational cost by reducing the problem dimension, but also facilitates a more detailed investigation of parametric bifurcations of the interface.

The derivations of effective interface equations are typically tedious and impose restrictions on the parameter regime for which they are valid~\cite{caginalp98_analy_phase_field_alloy_trans_layer, karma98_quant_phase_field_model_dendr, almgren99_secon_order_phase_field_asymp_unequal_conduc, elder2001sharp, boussinot13_inter_kinet_phase_field_model}.
In this article, we present a data-driven alternative for the identification of interface equations.
We illustrate our approach on an Allen–Cahn phase field model~\cite{allen72_groun_state_struc_order_binar, allen76_mechan_phase_trans_within_miscib, allen79_micros_theor_antip_bound_motion}, where an analytic derivation of approximate front partial differential equations (PDEs) describing the interface dynamics exists~\cite{elder2001sharp}.
One such interface equation is the Kardar–Parisi–Zhang (KPZ) equation, a low order approximation describing the interface dynamics in a relatively narrow parameter regime~\cite{kardar86_dynam_scalin_growin_inter}.
By learning the PDE in a data-driven way~\cite{gonzalez-garcia98_ident_distr_param_system, krischer93_model_ident_spatiot_varyin_catal_react, rico-martinez92_discr_vs,kemeth22_learn_emerg_partial_differ_equat},
we demonstrate the data-driven identification of an interface evolution equation that surpasses the accuracy of analytical interface models.

Note that there is a myriad of different approaches on how to identify the right-hand side of a PDE from data, such as sparse identification of nonlinear dynamical systems using dictionaries~\cite{brunton16_discov_gover_equat_from_data, rudy17_data_driven_discov_partial_differ_equat}, PDE-net~\cite{long17_pde_net}, and physics-informed neural networks~\cite{raissi19_physic_infor_neural_networ}, among others.
Here, we choose to represent the right hand side of the PDE operator (the ``law of the PDE'') through a neural network~\cite{gonzalez-garcia98_ident_distr_param_system, krischer93_model_ident_spatiot_varyin_catal_react, rico-martinez92_discr_vs,kemeth22_learn_emerg_partial_differ_equat}.

Finally, we exploit the analytically available closed form equations (such as the KPZ mentioned above) beyond their region of validity, and showcase the data-driven learning of corrections that rectify their predictions. Making use of (and correcting) the already derived physics is important in enhancing the interpretability of the models resulting from data-driven system identification.

\section{Interface models for phase field systems}
\begin{figure}[ht]
\centering
\includegraphics[width=\columnwidth]{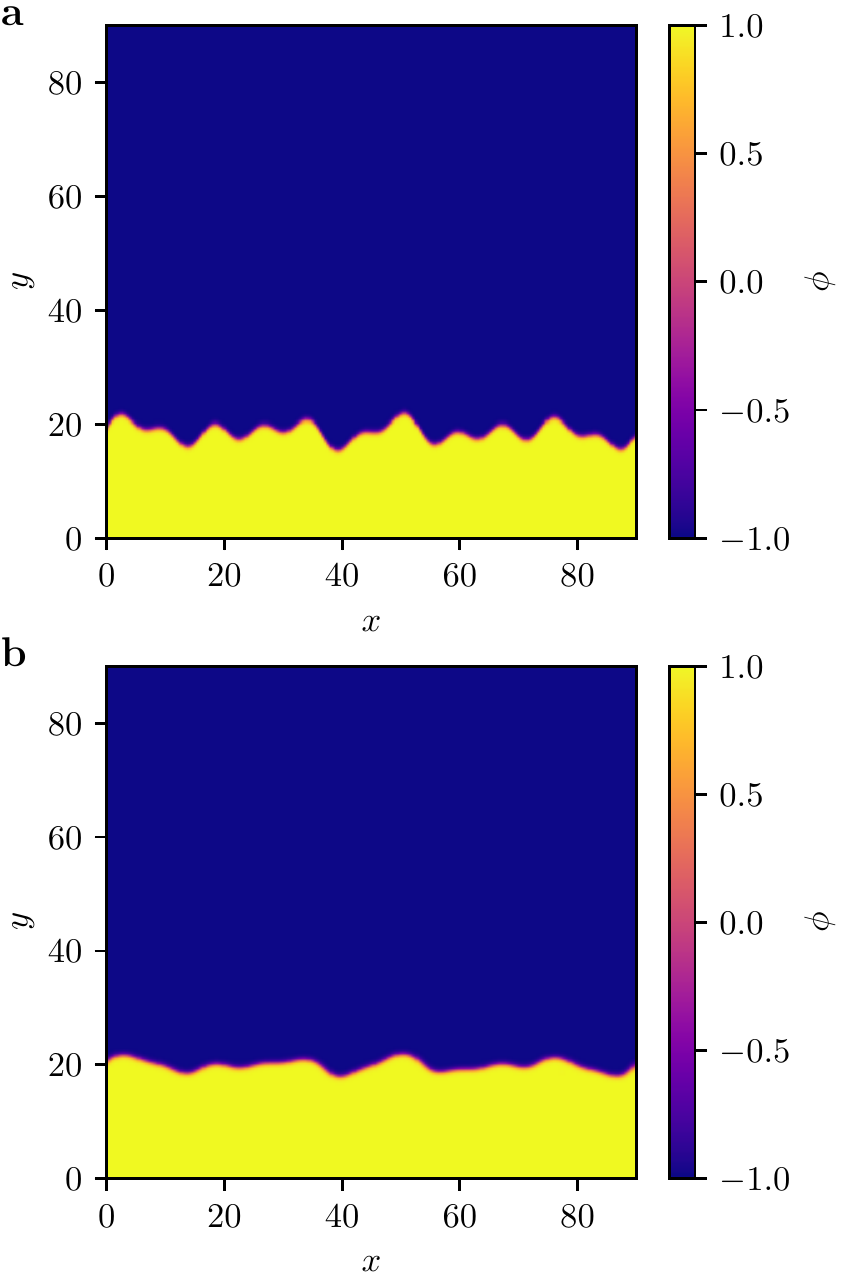}
\caption{(a) Initial profile of the phase field on a square domain of length $L=90$ with a resolution of $400\times 400$ pixels. (b) Profile after integrating the snapshot in (a) for $T=25$ dimensionless time units using the phase field system Eq.~\eqref{eq:pf} with $a=-0.1$ and $D=0.1$.
The color corresponds to the phase $\phi$.}
\label{fig:figure_1}
\end{figure}

We illustrate our approach on the phase field system described by
\begin{equation}
  \frac{\partial \phi}{\partial t} = D\nabla^2 \phi - \left(\phi-a\right)\left(\phi^2-1\right)
  \label{eq:pf}
\end{equation}
with the phase $\phi \in \left[-1, 1\right]$ and the parameters $a=-0.1$ and $D=0.1$.
This equation is integrated numerically on a two-dimensional square domain of length $L=90$ with $400$ pixels in each direction, see Methods.
A representative initial snapshot is shown in Fig.~\ref{fig:figure_1}(a) where the color encodes the phase field variable $\phi$.
As boundary conditions, we use zero-flux boundaries $\partial \phi/\partial y=0$ at $y=0$, $y=L$ and periodic boundaries $\phi\left(x=0\right)=\phi\left(x=L\right)$ at the left and right boundary.

\begin{figure}[ht]
\centering
\includegraphics[width=\columnwidth]{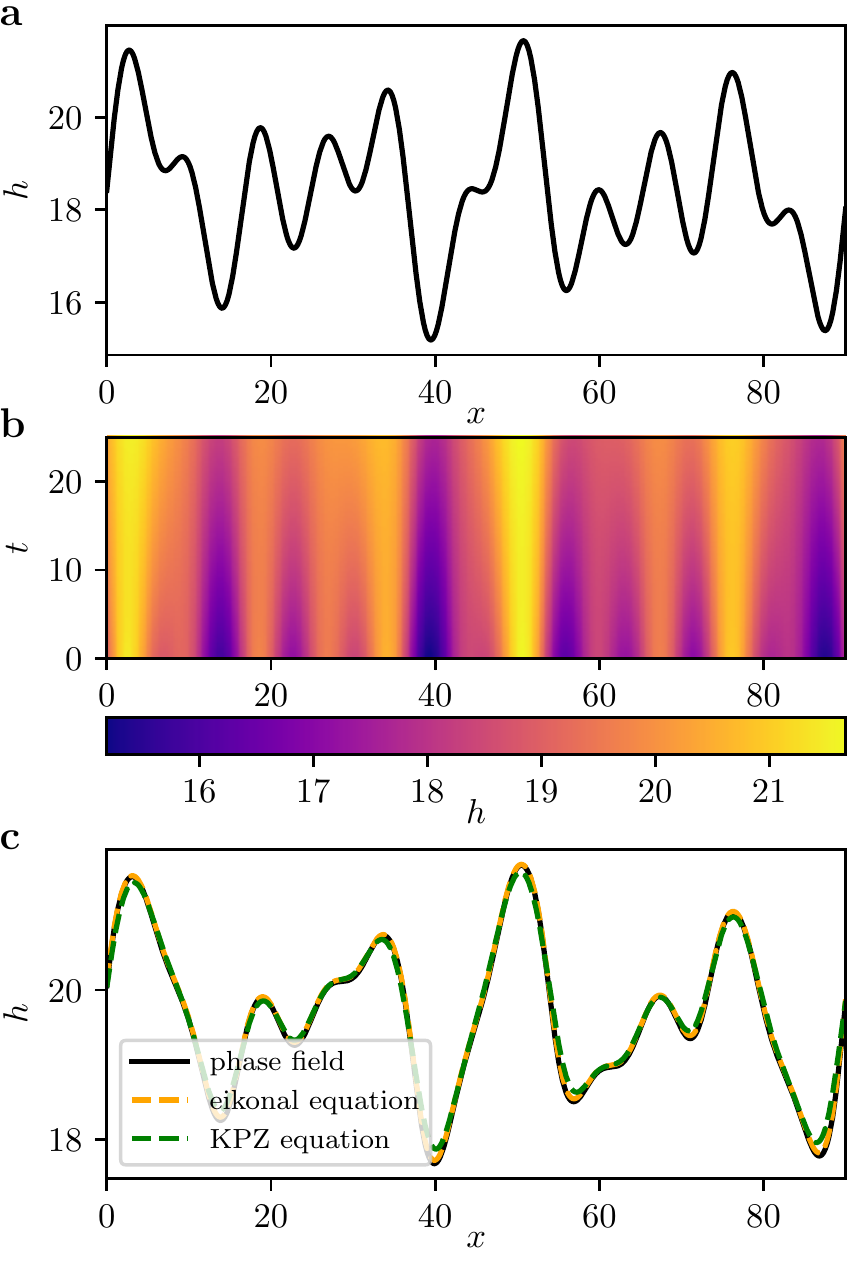}
\caption{(a) Initial front position obtained from the snapshot shown in Fig.~\ref{fig:figure_1}(a) by fitting the function Eq.~\eqref{eq:tanh} to the data, as explained in the main text.
  (b) Space-time dynamics of the front position, obtained from simulations of the phase field system Eq.~\eqref{eq:pf}.
  (c) Front position after integrating for $T=25$ dimensionless time units using the phase field system and extracting the front (black curve), using the eikonal equation, Eq.~\eqref{eq:front} (dashed orange curve), and using the KPZ equation, Eq.~\eqref{eq:kpz} (dashed green curve).}
\label{fig:figure_2}
\end{figure}

The snapshot resulting from numerically integrating Eq.~\eqref{eq:pf} for $T=25$ dimensionless time units is shown in Fig.~\ref{fig:figure_1}(b).
There, one can visually observe how the governing phase field system leads to a smoothening of the interface between the two phases at $\phi=1$ and $\phi=-1$.
This is better visualized by tracking the $y$-position of the phase front.
In order to achieve sharp interface contrast,
we obtain the $y$-positions, or height $h$, of the fronts from the simulation data by fitting
\begin{equation}
  f(y) = \tanh(cy-d)
  \label{eq:tanh}
\end{equation}
to the data at each time step and at each of the $400$ discrete $x$-positions.
The front position is then defined as $h = d/c$, that is where the $f(y)$ function crosses $\phi=0$.
The height $h$ of the front obtained this way for the initial snapshot shown in Fig.~\ref{fig:figure_1}(a) is plotted in Fig.~\ref{fig:figure_2}(a), whereas the front position of the snapshot at $T=25$ shown in Fig.~\ref{fig:figure_1}(b) is depicted in Fig.~\ref{fig:figure_2}(c) as a black curve.
In Fig.~\ref{fig:figure_2}(b), the space-time profile of the front position, obtained from the phase field solution, is shown, where the color encodes the front height $h$.

For the phase field example considered here, one can actually derive effective equations for the dynamics of the front position $h$ \cite{elder2001sharp}. First, assuming that the variations in the direction transverse to the motion are slower than in the longitudinal direction, at the sharp interface limit~\cite{elder2001sharp} one can obtain the eikonal equation for the normal velocity of the interface:
\begin{equation}
  v_n = -\sqrt{2D}a + D\kappa,
  \label{eq:interface}
\end{equation}
with $\kappa$ and $v_n$ given by
\begin{equation}
\kappa = \frac{1}{\left(1+\left(\frac{\partial h}{\partial x}\right)^2\right)^{3/2}} \frac{\partial^2 h}{\partial x^2},~~
v_n = \frac{1}{\sqrt{1+\left(\frac{\partial h}{\partial x}\right)^2}} \frac{\partial h}{\partial t}
\end{equation}
when written in Cartesian coordinates $(x,h(x))$.
This, in turn, means that the dynamics of the height $h$ of the front in the phase field
model Eq.~\eqref{eq:pf} discussed above can be described by the equation
\begin{equation}
  \begin{aligned}
    \frac{\partial h}{\partial t} &= \frac{D}{1+\left(\frac{\partial h}{\partial x}\right)^2}\frac{\partial^2 h}{\partial x^2} - \sqrt{2D}a\sqrt{1+\frac{1}{2}\left(\frac{\partial h}{\partial x}\right)^2}\\
    & = f_{eik}\left(h, \partial h/\partial x, \partial^2 h/\partial x^2\right) \, .
  \end{aligned}
  \label{eq:front}
\end{equation}
If, in addition, $\left|\partial h/\partial x\right| \ll 1$, then the dynamics can further be approximated by Taylor expanding the right hand side of Eq. (\ref{eq:front}), which, after disregarding cubic and higher-order terms, leads to the deterministic Kardar–Parisi–Zhang (KPZ) equation
\begin{equation}
  \begin{aligned}
    \frac{\partial \tilde{h}}{\partial t} &= D\frac{\partial^2 \tilde{h}}{\partial x^2} - a\sqrt{\frac{D}{2}}\left(\frac{\partial \tilde{h}}{\partial x}\right)^2\\
    & = f_{KPZ}\left(\tilde{h}, \partial \tilde{h}/\partial x, \partial^2 \tilde{h}/\partial x^2\right)
  \end{aligned}
  \label{eq:kpz}
\end{equation}
with $\tilde{h} = h+\sqrt{2D}at$.

We can now compare the dynamics of the front obtained from the phase field model, Eq.~\eqref{eq:pf},
to the dynamics of the front predicted by the eikonal and KPZ equations.
The predictions of the evolution of the initial front profile shown in Fig.~\ref{fig:figure_2}(a) using the two interface models and periodic boundary conditions
are plotted in Fig.~\ref{fig:figure_2}(c) as a dashed orange curve and dashed green curve, respectively.
In addition the ``ground truth'' front position obtained by integrating the two-dimensional phase field system and subsequently extracting the front position is shown as a solid black curve.
Note that there is a very close correspondence between the front position at $T=25$ obtained from the phase field model and obtained by integrating the eikonal model, Eq.~\eqref{eq:front}.
The predictions of the KPZ deviate slightly in places where the curvature of the interface is large.

\section{Black Box Front Dynamics}

\begin{figure}
  \centering
  \includegraphics[width=\columnwidth]{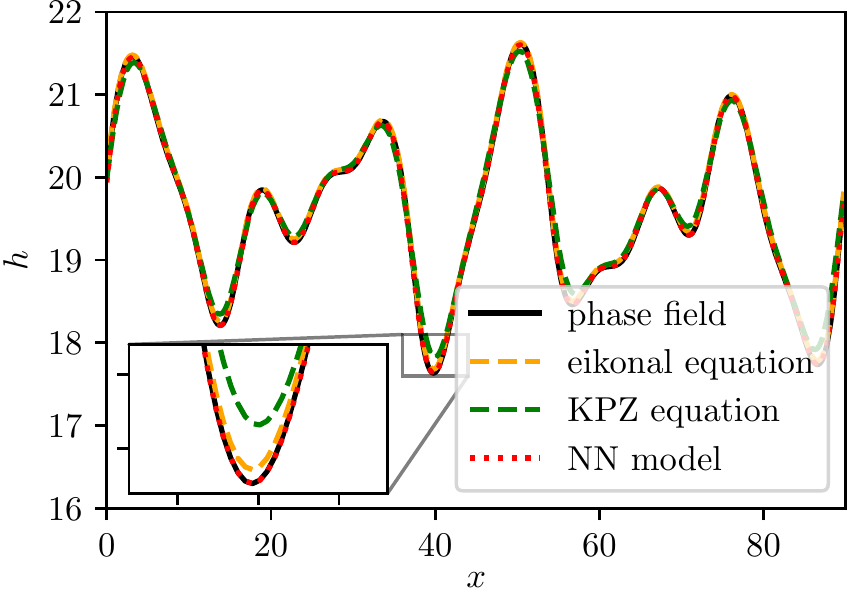}
  \caption{The front position at $T=25$ obtained by integrating the initial condition shown in Fig.~\ref{fig:figure_1}(a) using the phase field system and extracting the front (solid black curve),
    integrating the corresponding front as shown in Fig.~\ref{fig:figure_2}(a) using the eikonal equation, Eq.~\eqref{eq:front} (dashed orange curve), using the KPZ equation, Eq.~\eqref{eq:kpz} (dashed green curve), and using the learned PDE model, Eq.~\eqref{eq:nn_pde} (dotted red curve). The zoom in shows an enlargement of parts of the front position, with the absolute errors shown in Fig.~\ref{fig:figure_4}.
  Parameters are $a=-0.1$ and $D=0.1$.}
  \label{fig:figure_3}
\end{figure}

For many applications, and in broad parameter regimes, an analytic reduction to a reduced model of the front dynamics might not be possible.
In such cases, we can learn the dynamics of the front in a data-driven way, for example by
learning a partial differential equation represented by a neural network~\cite{gonzalez-garcia98_ident_distr_param_system, krischer93_model_ident_spatiot_varyin_catal_react, rico-martinez92_discr_vs,kemeth22_learn_emerg_partial_differ_equat}.
We do this here by integrating $N_{\mathrm{train}}=20$ different initial conditions in the two-dimensional
space domain, and subsequently extracting the front position as described above.
This gives $20$ space-time trajectories of $h$ on a one-dimensional spatial domain.
We subsequently calculate the time derivative of the front, $\partial h/\partial t$,
and the space derivatives, $\partial h/\partial x$ and $\partial^2 h/\partial x^2$,
at each point in space and time for each trajectory using finite differences.
We then use a fully connected neural network $NN_\Theta$ with weights $\Theta$ to represent the function
\begin{equation}
  \widehat{\frac{\partial h}{\partial t}} = NN_{\Theta}\left(h, \partial h/\partial x, \partial^2 h/\partial x^2\right).
  \label{eq:nn_pde}
\end{equation}
The weights are optimized by minimizing the mean-squared error between the output
of the neural network, $\widehat{\partial h/\partial t}$, and
the actual time derivatives of the $h$ field, see the Methods section for details on the training.
After training this neural network,
one obtains a partial differential equation, which, given proper initial conditions and boundary conditions, one can employ for simulation.
The predictions of this one-dimensional front neural network model using periodic boundary conditions,
as well as the predictions of the analytical PDEs, by integrating the initial front
given in Fig.~\ref{fig:figure_2}(a), are shown in Fig.~\ref{fig:figure_3}.
Note that there is very good agreement between the ground truth front dynamics (solid black curve) and the data-driven PDE (dotted red curve); However, due to the analytical approximations and the finite values of $a$ and $D$, the closed form eikonal and KPZ PDE predictions deviate slightly.

\begin{figure*}[ht]
  \centering
  \includegraphics[width=0.95\textwidth]{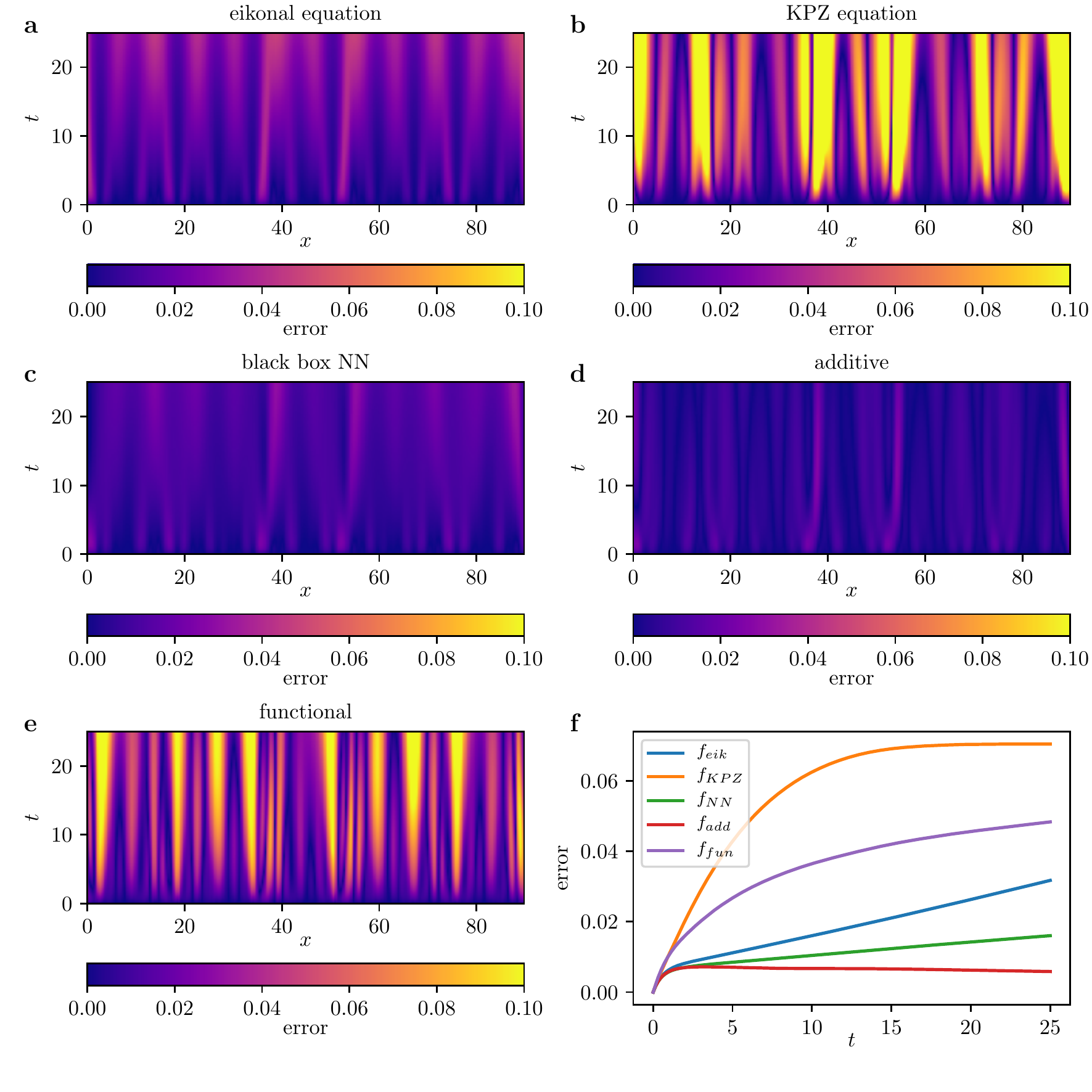}
  \caption{Absolute error at each point in space and time between the actual front position obtained from the phase field model (the test data, see Methods) and the predictions obtained by integrating the initial front snapshot from the test data using the different modeling approaches. 
  (a)~Absolute error when integrating using the analytical eikonal equation, Eq.~\eqref{eq:front}.
  (b)~Absolute error when integrating using the analytical KPZ equation, Eq.~\eqref{eq:kpz}.
  (c)~Absolute error when integrating using the data-driven black box PDE model, Eq.~\eqref{eq:nn_pde}.
  (d)~Absolute error when integrating using the additive gray-box PDE model, Eq.~\eqref{eq:gray_nn_pde}.
  (e)~Absolute error when integrating using the functional gray-box PDE model, Eq.~\eqref{eq:gray_nn_pde_fun}.
  (f)~Spatially-averaged absolute error over time of the different modeling approaches.
  For integration, SciPy's implementation of a Runge-Kutta 4(5) stepper~\cite{2020SciPy-NMeth, dormand80_famil_embed_runge_kutta_formul} was used.
    }
  \label{fig:figure_4}
\end{figure*}

This becomes more obvious when visualizing the absolute difference between the ground truth front position obtained from the phase field and the predictions of the surrogate models over space and time.
For the eikonal equation, Eq.~\eqref{eq:front}, the absolute difference is shown in Fig.~\ref{fig:figure_4}(a), with the absolute prediction error of the KPZ equation, Eq.~\eqref{eq:kpz}, presented in Fig.~\ref{fig:figure_4}(b), where yellow indicates large deviations.
In contrast, the error of the predictions obtained from the learned ``black box'' neural network PDE remain small over the time interval considered, see Fig.~\ref{fig:figure_4}(c).

\section{Gray Box Front Dynamics}

In the previous section, we learned a neural network PDE description that,
given the interface height $h$ as well as the spatial partial derivatives of $h$, predicts the time derivative $\partial h/\partial t$
using a large set of learned weights and nonlinear activation functions.
This makes such a PDE useful for prediction, yet barely accessible for interpretation, meaning that physical processes responsible for the dynamics, such as advection and diffusion, cannot be easily observed / disentangled from the resulting PDE model.
In contrast, having an approximate model at hand allows to include interpretable parts into the nonlinear system identification.
In particular, instead of learning a ``black box'' PDE model using a neural network as in the previous section, one may use a neural network to learn {\em the correction to the approximate model}.
We therefore phrase our first ``physics infused'' modification of the system identification task as finding a neural network that, when added to the approximate ``white box'' model, gives the correct time derivative of $h$ at each point in space and time.
That is, given the KPZ equation, Eq.~\eqref{eq:kpz}, we learn a neural network such that
\begin{equation}
  \begin{aligned}
    \widehat{\frac{\partial h}{\partial t}} & = f_{KPZ}\left(h, \partial h/\partial x, \partial^2 h/\partial x^2\right)\\
    & + NN_{\Theta}\left(h, \partial h/\partial x, \partial^2 h/\partial x^2\right)\\
    & = f_{add}\left(h, \partial h/\partial x, \partial^2 h/\partial x^2\right).
  \end{aligned}
  \label{eq:gray_nn_pde}
\end{equation}
As in the previous examples, we use a set of $20$ training trajectories, and optimize the model by minimizing the mean squared error between its output and $\widehat{\frac{\partial h}{\partial t}}- f_{KPZ}\left(h, \partial h/\partial x, \partial^2 h/\partial x^2\right)$.
We then evaluate the efficacy of this model on the initial condition shown in Fig.~\ref{fig:figure_2}(a). The absolute error over space and time between the integration results of the learned gray box model and the front position obtained from the original phase field system is shown in Fig.~\ref{fig:figure_4}(d).
Note that the error remains very small over the time window considered, with comparable magnitude to the black box predictions shown in Fig.~\ref{fig:figure_4}(c).

We now propose an alternative, second ``physics infused'' modification by exploiting our knowledge of the (now inaccurate) analytical closed form equation models.
We consider a (functional) correction of the form

\begin{equation}
  \begin{aligned}
    \widehat{\frac{\partial h}{\partial t}} & = NN_{\Theta}\left(f_{KPZ}, \partial f_{KPZ}/\partial x, \partial^2 f_{KPZ}/\partial x^2\right)\\
    & = f_{fun}\left(f_{KPZ}, \partial f_{KPZ}/\partial x, \partial^2 f_{KPZ}/\partial x^2\right).
  \end{aligned}
  \label{eq:gray_nn_pde_fun}
\end{equation}

This means that we learn the dynamics based on the local values and spatial derivatives of the KPZ closure only. 
Including higher derivatives of the analytical closure in the model (an alternative possibility would be to
include values of the analytical closure at nearby points) can be thought of as a spatial analog of the Takens embedding for dynamical systems attractor reconstruction~\cite{Whitney1936, Takens1981}.
In the Takens embedding, the effect of important missing variables is modeled through inclusion of short temporal histories (time delays) of the variables we can measure.
In our work here, the effect of important information missed by the analytical closure is modeled through the inclusion of short spatial histories / neighboring spatial profiles / ``space delays'' of the analytical closure values: we are, in a particular sense, guided by the analytical approximate closure, making the equation ``higher order in space''.

The prediction results 
of this ``functional'' gray box model is shown in Fig.~\ref{fig:figure_4}(e),
which indicates a slightly inferior performance than using a simple additive correction to the KPZ.

\section{Discussion}

We discussed -- and demonstrated -- the possibility to learn effective partial differential equations for the interface dynamics of phase field systems. This allows the reduction of the dimension of the problem at hand, facilitating tasks such as prediction or bifurcation analysis.

Our illustrative example was a model phase field system, where analytic reductions to the phase field dynamics exist.
In particular, an eikonal equation, Eq.~\eqref{eq:front}, and an approximation thereof, the KPZ equation, Eq.~\eqref{eq:kpz},
have been derived.
Here, we show that learning an interface PDE for the front dynamics based on simulation data, and parametrized by a fully-connected neural network, can lead to enhanced prediction accuracy.
Having access to approximate models, we furthermore highlight the ability to find {\em data-driven corrections} to such equations, as we demonstrate for the example of the KPZ equation.
We discuss two different ways to correct the operator of the approximate model: one can either learn an additive term that rectifies the output of the KPZ equation, or one can learn a correction based on the output of the KPZ equation sampled in a small spatial neighbourhood of the current point of interest.
Exploiting this partial physical knowledge renders the nonlinear system identification task  a {\em gray box} one (to be contrasted with the physics-uninformed black box version we started with). This enhances our understanding/interpretation~\cite{molnar2020interpretable} of the predictions of
the learned dynamical system: The white part of the gray box models corresponds to the KPZ equation,
while only the higher-order corrections are being learned by the black box neural network,
eventually making the model more easily accessible to human interpretation.

The performance of the different models discussed here is summarized in the mean prediction error over time as shown in Fig.~\ref{fig:figure_4}(f). 
The KPZ shows, as expected, an inferior performance than the eikonal equation, from which it is derived via supplementary approximations.
In contrast, black and gray box neural network models show superior or comparable prediction accuracy to the eikonal equation.

Different aspects, however, can influence the performance of the learned interface PDEs: 
\begin{itemize}
    \item The generation of the training data involves fitting the front position as well as estimating time and space derivatives (e.g. using finite differences, or spectral interpolation). These steps impose numerical inaccuracies which may deteriorate the performance of the neural network PDEs.
    \item The performance of neural networks in general depends on the choice of hyperparameters such as learning rate during training, the number of hidden layers, the optimizer, and, as in our case, the number of partial derivatives used as input to the models.
\end{itemize}

A thorough investigation of these different factors might further increase the accuracy of the proposed models, and we are currently working on a more detailed study. 
We are also exploring black and gray box applications to other types of amplitude equations (like the Kuramoto-Sivashinsky equation) and modulational envelope equations (like those arising in the continuum limit of the Fermi-Pasta-Ulam-Tsingou model) with special focus on when analytically available approximations begin to drastically fail. 

In the model phase field example considered, obtaining the phase field interface was comparatively simple (here, we fitted a tanh function at each $x$ point in space).
For other examples, estimating the front position (and its movement) is a more involved, nontrivial task.
Examples include the movement of cells, where the tracking of the cell boundary is a problem that attracts increasing research efforts~\cite{schindler2020analysis}.
We are currently working on finding surrogate interface PDE models (and stochastic SPDE models~\cite{Dietrich2021}) for such problems, in particular with respect to biological applications, such as cell deformation and migration processes during morphogenesis, wound healing, and cancer metastasis.

\section{Methods}
\label{sec:methods}

\subsection{Neural Network Training}

All neural networks presented in this article are composed of 4 layers with 96 neurons in each layer.
Each hidden layer is followed by a Swish nonlinear activation function~\cite{ramachandran17_swish}.
The neural networks take as input $h$ as well as the first two spatial derivatives $\partial h/\partial x$ and $\partial^2 h/\partial x^2$.
The spatial derivatives are calculated using a finite difference stencil of length \(l=5\) and
the respective finite difference kernel for
each spatial derivative of the highest accuracy order that fits into \(l=5\).
As the weight initializer and the other hyperparameters we take the default settings from PyTorch~\cite{PyTorch}.
The weights are optimized using the mean squared error between the network output and the objective using the Adam optimizer~\cite{kingma2017adam}.
As training data, we sample $20$ different initial conditions, each initial condition composed of a superposition of four sinus modes with random amplitudes, taken uniformly from the interval $\left[0, 1\right]$ and discrete frequencies, taken randomly from $2\pi/L \cdot \left\{0, 1, \dots, 31, 32\right\}$.
This enforces profiles that are periodic in $x$, and offer a large variety.
Finally, an offset drawn from the interval $\left[10, 20\right]$ was added to the created profiles.

To generate training data, the initial profiles $h$ created this way are mapped to two-dimensional phase field profiles $\phi$.
This is done by defining the phase as
\begin{equation}
\phi = \tanh\left(\left(h-y\right)/\sqrt{2D}\right).
\end{equation}
Finally, these $20$ phase profiles are integrated forward using the phase field model for $T=25$ dimensionless time units using SciPy's implementation of a Runge-Kutta 4(5) stepper~\cite{2020SciPy-NMeth, dormand80_famil_embed_runge_kutta_formul}.
At $500$ equidistant time steps between $0$ and $T$, we calculate the front position as described in the main text.
Finally, we calculate the space and time derivatives at each point in space and time of these 20 trajectories, resulting in the training data for the PDE models.
In addition, we test our models on the additional trajectory shown in Fig.~\ref{fig:figure_1} which is not contained in the training data, leading to a total of 20 training trajectories and 1 test trajectory.

\subsection*{Acknowledgment}
The work of FPK and IGK was partially supported by the US Department of Energy, and the US Air Force Office of Scientific Research. 
TM and CB acknowledge financial support by the Deutsche Forschungsgemeinschaft (DFG) -- Project-ID 318763901 -- SFB1294, project B02. BE acknowledges financial support by MICINN/AEI through research grant PID2020-116927RB-C22.

\bibliography{lit.bib}
\bibliographystyle{unsrt}

\end{document}